\documentclass[letterpaper, 10 pt, conference]{ieeeconf}
\IEEEoverridecommandlockouts

\usepackage{xspace}
\usepackage{xcolor}
\usepackage{amsfonts}
\usepackage{graphics} %
\graphicspath{{figures/}}  %
\usepackage{epsfig} %
\usepackage{mathptmx} %
\usepackage{times} %
\usepackage{amsmath} %
\usepackage{amssymb}  %
\usepackage{ulem}
\usepackage{siunitx}
\usepackage{verbatim}
\usepackage{booktabs}
\usepackage{multirow}
\usepackage[space]{cite}
\usepackage[table]{xcolor} %
\definecolor{lightgreen}{RGB}{210,255,210}
\definecolor{darkgreen}{RGB}{0,120,0}
\usepackage[colorlinks=true, linkcolor=blue, urlcolor=blue, citecolor=blue]{hyperref}
\usepackage{makecell}

\usepackage[font=footnotesize]{caption}

\makeatletter
\let\origthebibliography\thebibliography
\def\thebibliography#1{%
  \origthebibliography{#1}%
  \scriptsize %
}
\makeatother

\newcommand{\method}{StereoEngine\xspace}

\usepackage{pifont}
\usepackage{xcolor}
\usepackage{makecell}
\usepackage{graphicx}
\newcommand{\cmark}{\textcolor{green!55!black}{\ding{51}}}

\definecolor{figgray}{rgb}{0.2, 0.2, 0.2}
\definecolor{figgreen}{rgb}{0.180, 0.494, 0.196}
\definecolor{figblue}{rgb}{0.082, 0.396, 0.753}
\definecolor{figlightred}{rgb}{1.0, 0.149, 0.0}
\definecolor{figdarkblue}{rgb}{0.188, 0.392, 0.729}

\title{\LARGE \bf
\method: Scaling Stereo Bimanual Robot Data \\without Stereo Cameras
}

\author{
    Jason Chen
    \quad
    I-Chun (Arthur) Liu
    \quad
    Gaurav~S.~Sukhatme
    \quad
    Daniel Seita\\
    University of Southern California
}

\begin{document}
\twocolumn[{%
\renewcommand\twocolumn[1][]{#1}%
\maketitle
\begin{center}
    \centering
    \captionsetup{type=figure}
    \includegraphics[width=1\textwidth]{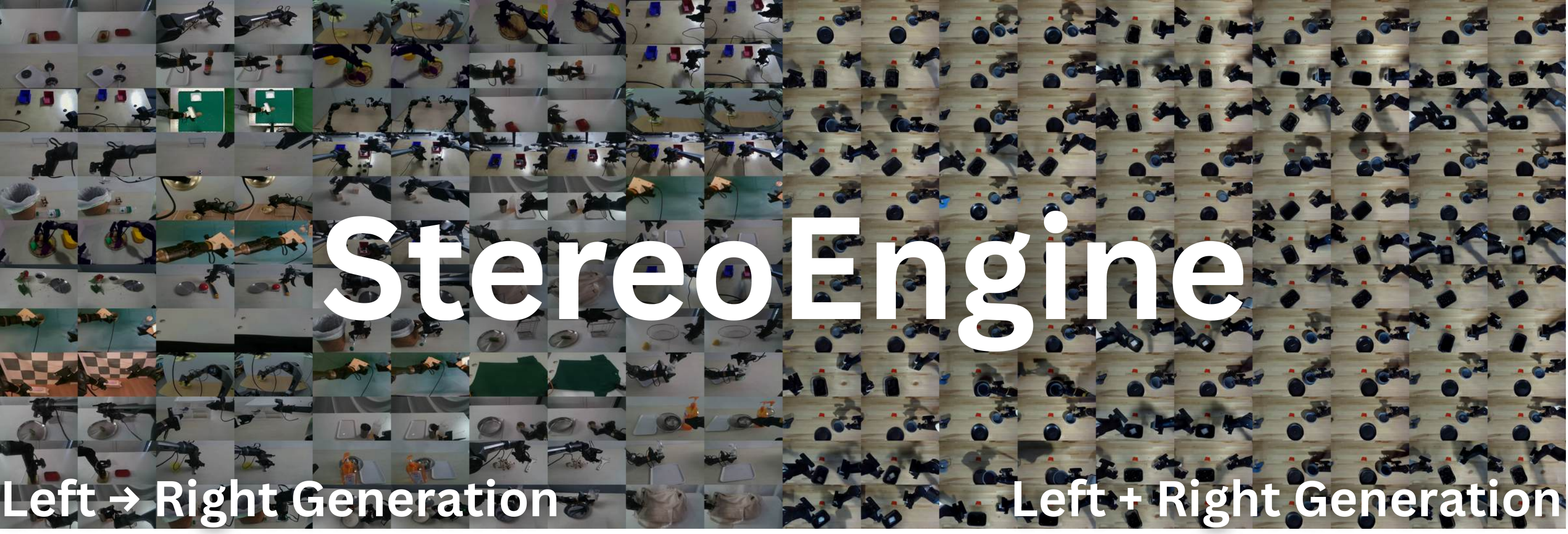}
    \vspace{-5pt}
    \caption{\textbf{\method generates stereo data for bimanual manipulation.} A mosaic of synthetic stereo pairs produced by our video diffusion framework where each adjacent column pair forms a left--right stereo view. The left half of the gallery shows our first method (\method-L) which takes a monocular real-world demonstration and synthesizes the corresponding right view to complete the stereo pair. The right half shows our second method (\method-S), which generates both left and right views jointly from simulated demonstrations, producing large-scale and photorealistic stereo data without any real video as input.}
    \label{fig:teaser}
\end{center}%
}]
\thispagestyle{empty}
\pagestyle{empty}

\begin{abstract}
Stereo vision is increasingly used for robotic manipulation with a growing body of work incorporating it into Vision-Language-Action (VLA) models. However, the scarcity of large-scale, real-world stereo datasets remains a fundamental bottleneck for scaling such policies. Recent generative monocular-to-stereo methods target settings such as VR content and autonomous driving. No prior work produces realistic synthetic stereo data for robotic manipulation where stereo supervision has come predominantly from simulation. We propose \method, which introduces two complementary methods to synthesize stereo data for robotic manipulation. In the first, we warp a monocular (left) video into the right view using estimated depth and inpaint the disoccluded regions with a video diffusion model which lifts monocular demonstrations into stereo pairs without training on real stereo. In the second, we render a stereo pair in simulation and jointly generate both photorealistic views which produces consistent left--right data without any real video as input. \method outperforms existing baselines in generated stereo quality, and using its synthetic data for policy mid-training or task-specific fine-tuning substantially improves downstream bimanual manipulation. Project website: \href{https://stereoengine.github.io}{https://stereoengine.github.io}.
\end{abstract}

\section{Introduction}
\begingroup
  \setlength{\skip\footins}{1pt}    %
  \setlength{\footnotesep}{2pt}     %
  \let\thefootnote\relax
  \footnotetext{%
    All authors are with the Thomas Lord Department of Computer Science at the University of Southern California, USA. Correspondence: \texttt{jchen567@usc.edu}.%
  }
\endgroup
Imitation learning from teleoperated demonstrations has enabled increasingly capable bimanual manipulation systems~\cite{aloha2_2024,intelligence2025pi05,intelligence2026pi07steerablegeneralistrobotic}. Recent stereo-conditioned policies extend this by taking stereo observations as input and exploiting binocular views to improve spatial reasoning, including challenging cases such as transparent objects~\cite{han2026stereopolicyimprovingroboticmanipulation,deng2026stereovlaenhancingvisionlanguageactionmodels}. However, scaling these stereo-conditioned policies to the point where they can be used for large-scale general imitation learning remains difficult, as large-scale robot datasets are overwhelmingly monocular and single-arm~\cite{open_x_embodiment_rt_x_2023,fang2026molmoact2actionreasoningmodels,wu2026robocoinopensourcedbimanualrobotic}. The few datasets captured with stereo cameras, such as DROID~\cite{khazatsky2024droid}, are single-arm which means no large-scale bimanual stereo dataset exists. Collecting stereo demonstrations at comparable scale would require expensive stereo rigs and costly re-teleoperation. This motivates scalable methods to synthesize stereo data without additional real-world collection~\cite{mandlekar2023mimicgen,xue2025demogen,yu2025real2render2realscalingrobotdata,chen2026craft}.

While monocular-to-stereo synthesis has emerged as a promising direction~\cite{zhao2024stereocrafter,dai2025svg,lu2026grt}, existing works target domains outside of robotic manipulation, such as VR/AR content~\cite{2024spatialdreamer,yu2025mono2stereobenchmarkempiricalstudy,qiao2025genstereo,xing2025stereoworldgeometryawaremonoculartostereovideo}. 
Reconstruction-based novel-view synthesis (e.g., NeRF~\cite{mildenhall2020nerf}, 3D Gaussian Splatting~\cite{kerbl3Dgaussians}) requires multi-view datasets and per-scene optimization on static scenes, unlike our dynamic single-view setting.
Methods that train on stereo pairs draw on simulation data that suffers from domain gaps or on only a small number of real pairs~\cite{huang2025restereodiffusionstereovideo}. To our knowledge, no prior work synthesizes realistic stereo data directly from real robot video data.

To address the data bottleneck, we present \method, a scalable framework for synthesizing photorealistic stereo data for bimanual manipulation using video diffusion. \method provides two independent methods. When a real monocular demonstration exists, we convert it into a stereo pair by warping the left view and inpaint the disoccluded regions which are the surfaces that become visible in the right view but were never observed in the left. This preserves the appearance of the real data, though every generated pair remains tied to an existing demonstration. When no real video exists, we instead generate both views from a simulated trajectory, which allows data generation for entirely new tasks and configurations. Together, the two methods lift the abundant monocular data that already exists and generate new data where none exists, without ever training on real stereo correspondences.

We evaluate \method across 480 real-world rollouts spanning five bimanual manipulation tasks. See Figure~\ref{fig:teaser} for a visualization of the diverse synthetic stereo pairs produced by \method. 
Using ExStereo~\cite{liu2027exstereo} and StereoVLA~\cite{deng2026stereovlaenhancingvisionlanguageactionmodels} as our downstream policies, \underline{mid-training}~\cite{liu2027exstereo} them on \method data---an intermediate stage between a policy's large-scale pre-training and task-specific fine-tuning---or fine-tuning them on it directly, consistently improves manipulation performance and generalization over baselines.

The contributions of \method are:
\begin{itemize}
\item \method-L, a monocular-to-stereo pipeline whose inpainter is trained without any real stereo pairs, using warp-derived masks, partial arm masks, and disparity dropout.
\item \method-S, which jointly generates both left and right views from simulation, needing no real video.
\item A large-scale bimanual robot stereo dataset released from both methods.
\item Real-world experiments showing that \method data improve stereo-conditioned policies through both large-scale mid-training (\method-L) and task-specific fine-tuning (\method-S).
\end{itemize}

\begin{figure*}[t]
    \centering
\includegraphics[width=1\textwidth]{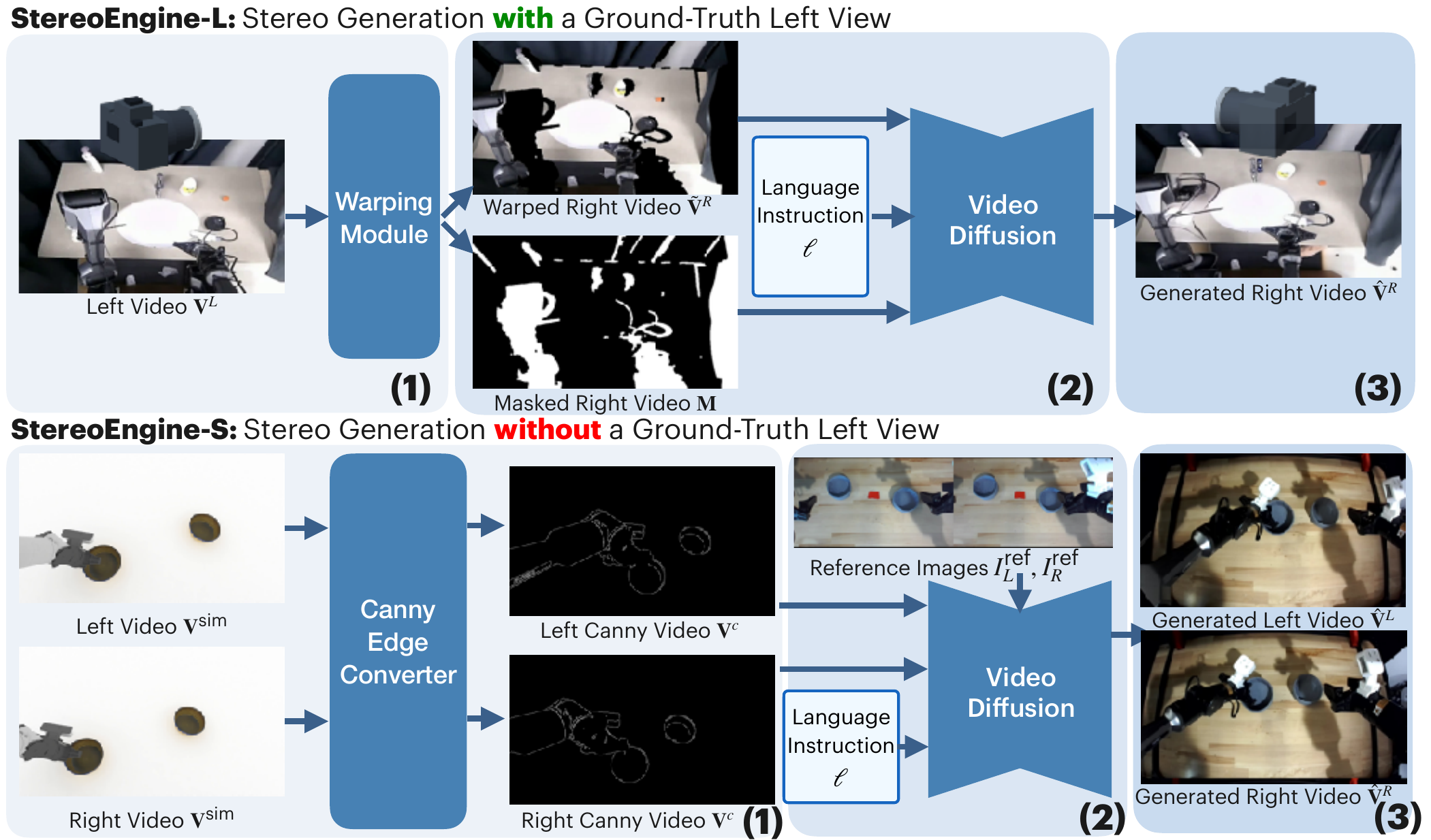}
    \caption{\textbf{Method Overview.} \textbf{\method-L} applies
    when a ground-truth left video is available. The warping module (1) moves every left pixel
    to its position in the right view, giving a warped right video and a mask marking the
    pixels no left pixel could reach. A video diffusion model (2) fills the masked region from
    these and a language prompt $\ell$, giving a right video (3) that pairs with the original
    left. \textbf{\method-S} applies when no real video exists. We render (1) the episode in
    simulation from two cameras, convert both renders to Canny edges, and (2) pass the edge videos
    to the model with two reference images and the instruction $\ell$. The edge map defines the scene geometry, while the reference images supply the appearance of the real workspace such as background, lighting, and surface texture. Then (3) a video generation model generates the synthetic video. In both methods, the resulting stereo pairs can be used to train a downstream bimanual stereo-conditioned policy.}
    \vspace{-10pt}
    \label{fig:pipeline}
\end{figure*}

\section{Related Work}

\subsection{Stereo Vision in Computer Vision}
\label{sec:related_stereo}
Stereo vision offers an efficient way of recovering 3D structure from image pairs where 
depth is obtained by estimating the disparity between a rectified left--right pair. This means obtaining
3D information from ordinary RGB cameras without active depth sensors. Recent stereo matching has shifted toward general-purpose foundation models with strong
zero-shot generalization, largely by leveraging priors from monocular depth foundation models
to generalize across unseen domains and remain reliable on transparent, reflective, and
textureless surfaces~\cite{wen2025stereo,Bartolomei_2025_CVPR,guo2024stereo}. These advances make stereo an accurate and deployable source of 3D
geometry well suited to robotics. In this work, we leverage these benefits for robot manipulation by generating large-scale stereo data so that stereo-conditioned policies can scale for training.

\subsection{Monocular-to-Stereo Generation}
Monocular-to-stereo generation synthesizes a right-eye view from a monocular left video, originally for immersive 3D content~\cite{xie2016deep3dfullyautomatic2dto3d}. Most methods require stereo pairs for training, either from large-scale stereo datasets that are expensive to obtain or from synthetic renders with a large domain gap from real video~\cite{zhao2024stereocrafter,yu2025mono2stereobenchmarkempiricalstudy,qiao2025genstereo,xing2025stereoworldgeometryawaremonoculartostereovideo}; training-free~\cite{wang2024stereodiffusion,dai2025svg,shi2026dissolvestereocoarsedepthinjection} and recent self-supervised~\cite{lu2026grt} variants relax this requirement. However, all of these only convert an existing left video and cannot produce stereo pairs where none exists. As summarized in Table~\ref{tab:comparison}, \method addresses both limitations: it needs no stereo pairs, operates on real robot data, and jointly generates both photorealistic views without any real footage.
\begin{table}[t]
\centering
\footnotesize
\setlength{\tabcolsep}{4pt}
\renewcommand{\arraystretch}{1.2}
\resizebox{\columnwidth}{!}{%
\begin{tabular}{llccccc}
\toprule
\textbf{Method} & \textbf{Domain}
& \makecell{\textbf{Video}\\\textbf{Gen.}}
& \makecell{\textbf{No Stereo}\\\textbf{Pairs}}
& \makecell{\textbf{Variable}\\\textbf{Baseline}}
& \makecell{\textbf{Both}\\\textbf{Views}}
& \makecell{\textbf{Gen.}\\\textbf{Length}} \\
\midrule
StereoCrafter~\cite{zhao2024stereocrafter}    & VR         & \cmark &        & \cmark &        & 25$^{*}$ \\
SpatialDreamer~\cite{2024spatialdreamer}      & VR         & \cmark & \cmark & \cmark &        & 8       \\
Mono2Stereo~\cite{yu2025mono2stereobenchmarkempiricalstudy} & 3D Content & & & & & 1 \\
GenStereo~\cite{qiao2025genstereo}            & XR/Driving &        &        & \cmark &        & 1       \\
StereoWorld~\cite{xing2025stereoworldgeometryawaremonoculartostereovideo} & XR & \cmark & & & & 81 \\
SVG~\cite{dai2025svg}                         & VR         & \cmark & \cmark & \cmark &        & 16      \\
Geo. Recip.~\cite{lu2026grt}                  & VR/AR      & \cmark & \cmark &        &        & --$^{\dagger}$      \\
\midrule
\textbf{\method (Ours)}                       & \textbf{Robotics} & \cmark & \cmark & \cmark & \cmark & \textbf{81} \\
\bottomrule
\end{tabular}%
}
\caption{Comparison of stereo and view-synthesis methods. A checkmark (\cmark) indicates the capability is demonstrated or explicitly supported by the method; a blank cell indicates it is not. \textbf{Video Gen.}: operates on videos instead of single images. \textbf{No Stereo Pairs}: requires no real stereo correspondences for training. \textbf{Variable Baseline}: synthesizes stereo videos with a controllable stereo baseline instead of a fixed one. \textbf{Both Views}: synthesizes both left and right views. \textbf{Gen. Length}: maximum number of frames generated in a single pass. $^{*}$Extendable to arbitrary length via autoregression. $^{\dagger}$No public code available and generation length not reported in the paper.}
\label{tab:comparison}
\vspace{-5pt}
\end{table}

\subsection{Stereo Vision in Robotics}
Stereo perception offers a practical route to 3D geometry in manipulation, recovering
structure on materials where active depth sensors fail, such as transparent, specular, and
textureless surfaces. Stereo can also yield cleaner geometry than active depth sensors, as we show later in Section~\ref{ssec:warping}. Prior work adapts stereo depth estimation to these challenging
conditions~\cite{wei2024droma,bai2024cleardepth} and uses stereo depth
for downstream grasping and tracking~\cite{yu2025pogs}. More recently, stereo has been integrated into imitation and Vision-Language-Action (VLA) policies. ActiveGlasses~\cite{zou2026activeglasseslearningmanipulationactive} reconstructs a point cloud from the stereo pair through a stereo matcher and feeds that to the policy, while StereoPolicy~\cite{han2026stereopolicyimprovingroboticmanipulation} and StereoVLA~\cite{deng2026stereovlaenhancingvisionlanguageactionmodels} consume the stereo images directly. All of these methods can benefit from large-scale stereo demonstrations, yet none address how such data can be obtained at scale. Stereo demonstrations can be teleoperated on a stereo rig, but this requires stereo hardware on every platform and teleoperation for every new task. \method instead generates the stereo data needed to train these policies without additional real-world collection.

\subsection{Video Generation in Robotics}
Video generation models have recently been used to scale robot data, either as world models that generate demonstrations for policy learning~\cite{jang2025dreamgen,gao2026dreamdojo,ye2026worldactionmodelszeroshot,nvidia2025cosmosworldfoundationmodel,ye2026robodream} or as data engines that augment existing demonstrations while keeping the robot trajectory fixed~\cite{yu2023scaling,robosplat,liu2025DCODA,chen2026ropa}. 
A related group of methods conditions video diffusion on simulation to synthesize new configurations or diverse demonstrations~\cite{mandlekar2023mimicgen,yu2025real2render2realscalingrobotdata}. 
Closest to our setting is CRAFT~\cite{chen2026craft}, which generates a video from Canny edges~\cite{canny1986edge} of a simulated bimanual episode together with a real reference image. 
However, all of these generate only monocular video, and \method is the first to extend video generation for robot data to stereo pair synthesis.

\section{Problem Statement}
\label{sec:problem}
We focus on scalable data generation for stereo-based policies in bimanual manipulation, where a policy $\pi_\theta$ with parameters $\theta$ is trained from stereo demonstrations. We write a camera's ground-truth left and right views at time $t$ as $I^{L}_{t}$ and $I^{R}_{t}$, and synthesized views with a hat (e.g., $\hat{I}^{R}_{t}$). We consider two independent settings for generating synthetic stereo pairs. In the first, a ground-truth left view $I^{L}_{t}$ is available and we synthesize the corresponding right view $\hat{I}^{R}_{t}$, forming the pair $(I^{L}_{t}, \hat{I}^{R}_{t})$. In the second, no ground-truth view is available and we generate both views jointly, $(\hat{I}^{L}_{t}, \hat{I}^{R}_{t})$. We present a method for each in Section~\ref{sec:method}. At deployment, the policy receives a real stereo pair $(I^{L}_{t}, I^{R}_{t})$
together with additional context $c_{t}$, and produces actions
$a_t = \pi_\theta(c_{t}, I^{L}_{t}, I^{R}_{t})$. The context
$c_{t} = \{s_{t}, \dots, \ell\}$ gathers whatever additional information the
policy has access to, such as the robot's proprioceptive state $s_{t}$ and a
language instruction $\ell$. In our experiments, $c_{t}$ contains only the
proprioceptive state $s_{t}$.

\section{Method: \method}
\label{sec:method}
\method addresses these two settings with two independent methods, each using a video diffusion model to synthesize photorealistic training videos (see Figure~\ref{fig:pipeline}).

\method-L targets the first setting. Given a ground-truth left video $\mathbf{V}^{L} = \{I^{L}_{1}, \ldots, I^{L}_{T}\}$ and a language instruction $\ell$, it synthesizes the corresponding right video $\hat{\mathbf{V}}^{R} = \{\hat{I}^{R}_{1}, \ldots, \hat{I}^{R}_{T}\}$ through two stages: warping (Section~\ref{ssec:warping}) and inpainting (Section~\ref{ssec:inpainting}). We describe how we train the inpainting model in Section~\ref{ssec:inpaint-training}.

\method-S targets the second setting. Given a simulation-rendered source video $\mathbf{V}^{\text{sim}} = \{I^{\text{sim}}_{1}, \ldots, I^{\text{sim}}_{T}\}$, reference images $I^{\text{ref}}$, and a language instruction $\ell$, it generates both photorealistic views jointly through two stages: simulation trajectory generation (Section~\ref{ssec:trajectory}) and video generation (Section~\ref{ssec:video-generation}). The instruction $\ell$ is different between the two methods.

\subsection{\method-L: Stereo Generation from a G.T. Left View}
\label{ssec:method1}
Following prior monocular-to-stereo work~\cite{xie2016deep3dfullyautomatic2dto3d,zhao2024stereocrafter,dai2025svg}, we treat the input monocular video as the ground truth (G.T.) left view and synthesize the corresponding right view. This choice of left to right generation instead of right to left is a convention in most stereo generation works and does not affect the method. Because no large-scale bimanual robotic stereo datasets are publicly available, naively training a model to synthesize right views is non-trivial. Our method generates right views \emph{without training a model on stereo pairs}. Unlike prior monocular-to-stereo methods, \method-L trains its inpainter only on unpaired monocular robot video, using warp-derived masks, partial arm masks, and disparity dropout to match inference and generalize across stereo baselines.

\begin{figure}[t]
    \centering
    \includegraphics[width=0.48\textwidth]{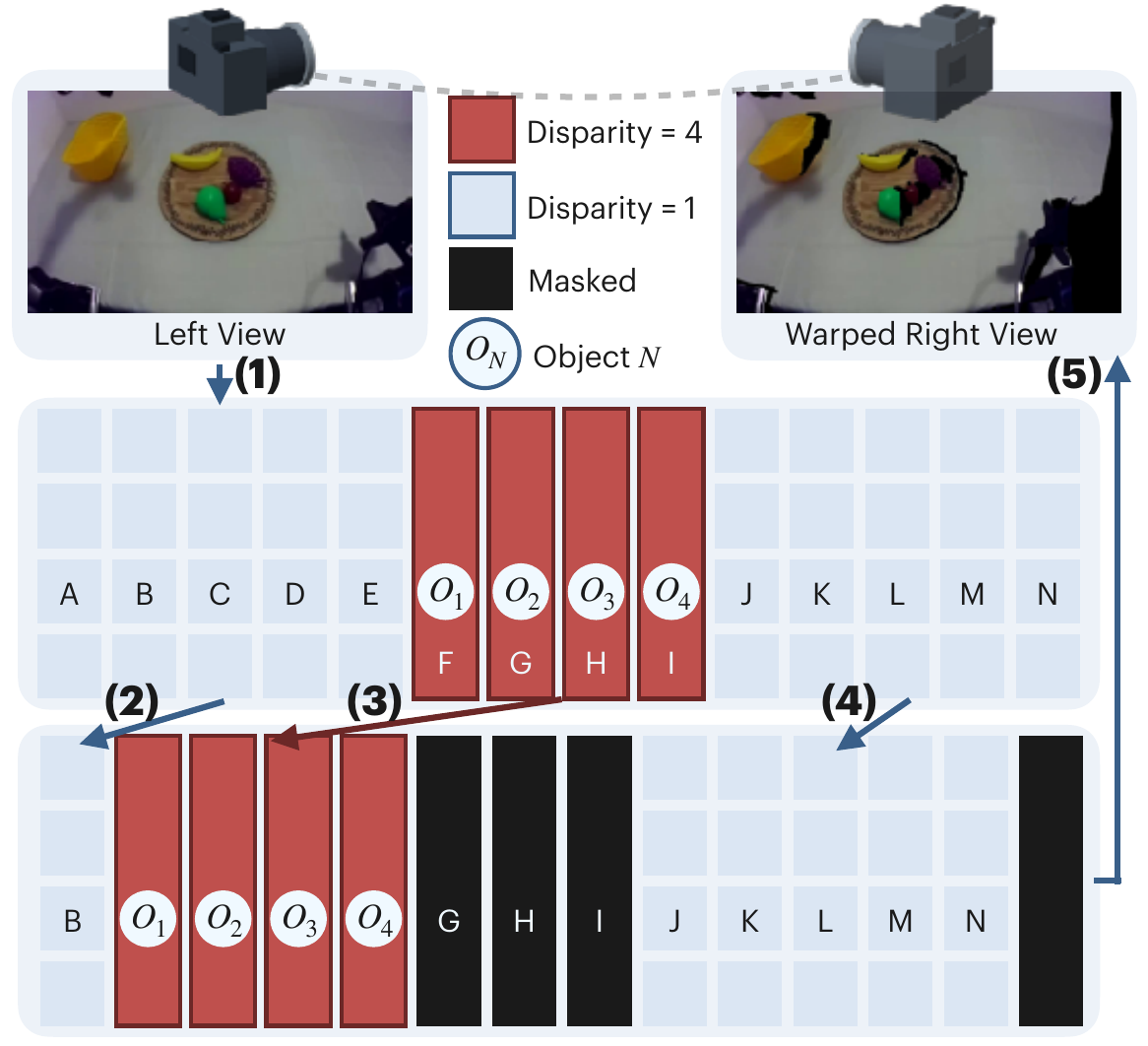}
    \caption{
        \textbf{Warping Module.}
        \textbf{(1)} the ground truth left view is abstracted into a grid of pixels, where $A$ to $N$ label the
        background and $O_1$ to $O_4$ label the object, and each pixel is given the disparity it must move by,
        $d=4$ for the object and $d=1$ for the background.
        \textbf{(2)} every pixel moves left by its own disparity, since the camera translates right. The object
        moves further and covers $C, D, E$, and the nearer pixel is kept, so only $B$ survives.
        \textbf{(3)} the background $G, H, I$ was hidden behind the object and never observed, so nothing can fill
        the gap it leaves and those columns become the mask.
        \textbf{(4)} the rest of the background shifts left by one, leaving the right border without any source.
        \textbf{(5)} the grid is mapped back to pixels to give the warped right view, correct where a pixel arrived
        and empty where the left camera never saw the surface.
    }
    \label{fig:warping_module}
    \vspace{-10pt}
\end{figure}

\subsubsection{Warping of the Left Video}
\label{ssec:warping}
Given the ground-truth left video $\mathbf{V}^{L}=\{I^{L}_{t}\}_{t=1}^{T}$, we feed $\mathbf{V}^{L}$ through a
monocular video depth model~\cite{video_depth_anything}, which outputs an  inverse depth map
$\tilde{d}_{t}$ for each frame. We place a virtual right camera with the same orientation and intrinsics by performing a horizontal translation of the left
camera. Figure~\ref{fig:warping_module} illustrates the procedure on
a grid of pixels.
Each pixel of the left frame corresponds to a 3D point at depth $Z$ and moving the camera sideways by a
baseline $B$ shifts that point horizontally by
\begin{equation}
d \;=\; \frac{fB}{Z} \quad \text{pixels},
\label{eq:disparity}
\end{equation}
where $f$ is the focal length in pixels and $d$ is the disparity. Since disparity is inversely proportional to depth, the predicted inverse depth $\tilde{d}$ gives the relative disparity across the image up to an unknown global scale. We fix this scale by setting $d = s\,\tilde{d}$ with $s = \delta_{0}/\operatorname{median}(\tilde{d})$, so that the median disparity equals a chosen $\delta_{0}$, which assigns each pixel the disparity it must move by (see Figure~\ref{fig:warping_module} stage~1). Any choice of $\delta_{0}$ produces a geometrically self-consistent pair, and since the downstream point cloud is reconstructed from the stereo pair, the absolute scale is never used. For \method, we set $\delta_{0}\!=\!63$ pixels ($\sim$7.6\% of frame width, comparable to
prior work~\cite{dai2025svg}) at deployment.

We leverage forward splatting~\cite{niklaus2020softmax} to map
each left frame pixel to its target position and splatting it onto the nearest right frame pixel. In the case that several pixels land on the same target, we weight each by its disparity using
$w=\sqrt{2}^{\,d}$ so the one with the largest disparity dominates and correctly occludes those behind it (see Figure~\ref{fig:warping_module} 
stage~2). However, this mapping does not cover the whole right frame. Surfaces hidden behind an object in the left view become
visible once the camera moves right which have no source pixel so we mark every empty target as occluded
which gives the disocclusion mask $\mathbf{M}$ (see Figure~\ref{fig:warping_module} stage~3). We dilate
$\mathbf{M}$ slightly so that it also covers the stretched pixels bordering large depth steps. A mask along the
right border is empty as the pixels that would fill it are outside of the left image
(see Figure~\ref{fig:warping_module} stage~4).

Warping therefore produces a right video $\tilde{\mathbf{V}}^{R}$ whose filled pixels are ground truth
observations from the left view, moved to the positions they occupy under the right camera, together with the mask $\mathbf{M}$ marking every pixel the left camera never 
observed, which the next stage fills (see Figure~\ref{fig:warping_module} stage~5).

\subsubsection{Inpainting of the Right Video}
\label{ssec:inpainting}
Diffusion-based video generative models learn to approximate a distribution
over video sequences through iterative denoising and the model learns the conditional distribution:
\begin{equation}
p_\psi(\hat{\mathbf{V}}^{R} \mid \tilde{\mathbf{V}}^{R}, \mathbf{M}, \ell),
\end{equation}
where $\tilde{\mathbf{V}}^{R}$ is the warped right video from Section~\ref{ssec:warping}, $\mathbf{M}$ is the
disocclusion mask, $\ell$ is a text prompt, and $\psi$ denotes
the model parameters. We use Wan2.2-VACE-Fun~\cite{wan2025} as our backbone. Wan2.2-VACE-Fun inpaints poorly zero-shot on this out-of-distribution robot domain, so we fine-tune it for our setting. We pass the warped video and the mask both as the region to be filled and as the
control video which tells the model exactly which pixels it may synthesize and which it must preserve.
The completed $\hat{\mathbf{V}}^{R}$ is
then paired with $\mathbf{V}^{L}$ to form the stereo pair.

\subsubsection{Training of the Inpainting Model}
\label{ssec:inpaint-training}
Since no large-scale robotic stereo dataset exists, the inpainting model $p_\psi$ from Section~\ref{ssec:inpainting} cannot be trained on real stereo pairs. It can
instead only be trained on ordinary monocular video since the ground truth for a synthetic mask is the clip itself.

We first pull 3,469 episodes of robot data of the same embodiment from public sources such as HuggingFace.
We deploy on AgileX PiPER arms in the real world so we collect datasets recorded with PiPER arms. We also
include the AgileX Cobot Magic platform which uses the same PiPER arms mounted on a mobile base with a leader
and follower system. None of these datasets contain stereo pairs. Our goal is to generate synthetic masks over these ground truth videos.

To generate a mask we apply the same forward-warp operation of Section~\ref{ssec:warping} to each of the clips
with the target median disparity $\delta_{0}$ resampled per clip so that the masks have the same shape and placement
distribution as they will have at inference compared to rectangles or free-form strokes used by generic
inpainting datasets. We sample $\delta_{0}$ uniformly between 35 and 95 pixels per clip when building training data to make the model capable of inpainting any baseline.

One failure mode of training on these masks alone is that it learns to inpaint the background and leave the
foreground untouched. To teach the model to inpaint the arm, we therefore add partial arm masks that cover a subset of the
manipulator. We also apply random disparity dropout to 10\% of clips and sample a dropout rate uniformly between 0.05 and 0.25.

\begin{table*}[t]
\centering
\footnotesize
\setlength\tabcolsep{3pt}
\begin{tabular}[t]{@{}c@{}}
\textbf{\method-L: Left $\rightarrow$ Right Generation}\\[3pt]
\begin{tabular}{lccc >{\columncolor{lightgreen}}c}
\toprule
\textbf{Base Policy} & \multicolumn{4}{c}{\textbf{ExStereo}} \\
\cmidrule(lr){2-5}
    Task & \makecell{No\\Mid.} & SVG & \makecell{Stereo\\Crafter} & \textbf{\method} \\
\midrule
Stack Two Bowls     & 9 / 20  & 7 / 20  & 10 / 20 & \textbf{13 / 20} \\
Lift Transparent Bottles & 11 / 20 & 10 / 20 & 13 / 20 & \textbf{18 / 20} \\
Close Door  & 16 / 20 & 14 / 20 & 17 / 20 & \textbf{19 / 20} \\
\bottomrule
\end{tabular}
\end{tabular}
\hspace{2em}
\begin{tabular}[t]{@{}c@{}}
\textbf{\method-S: Left + Right Generation}\\[3pt]
\begin{tabular}{l c>{\columncolor{lightgreen}}c c>{\columncolor{lightgreen}}c}
\toprule
\textbf{Base Policy} & \multicolumn{2}{c}{\textbf{ExStereo}} & \multicolumn{2}{c}{\textbf{StereoVLA}} \\
\cmidrule(lr){2-3}\cmidrule(lr){4-5}
    Task
    & \makecell{Point\\Clouds} & \textbf{\method}
    & \makecell{Point\\Clouds} & \textbf{\method} \\
\midrule
Stack Two Bowls  & 0 / 20 & \textbf{11 / 20} & 0 / 20 & \textbf{7 / 20} \\
Lift Two Bottles & 0 / 20 & \textbf{13 / 20} & 0 / 20 & \textbf{9 / 20} \\
Container Plate  & 0 / 20 & \textbf{15 / 20} & 0 / 20 & \textbf{11 / 20} \\
\bottomrule
\end{tabular}
\end{tabular}
\caption{\textbf{Real-world results.} Success rate over 20 trials (higher is better). The \textbf{Base Policy} is the stereo-conditioned policy being trained (ExStereo~\cite{liu2027exstereo} or StereoVLA~\cite{deng2026stereovlaenhancingvisionlanguageactionmodels}), and each column is a different training condition. In \textbf{\method-L}, each column is a different mid-training condition: \textbf{No Mid.} trains the policy directly with no stereo mid-training, and the remaining columns mid-train it on stereo data from SVG~\cite{dai2025svg} and StereoCrafter~\cite{zhao2024stereocrafter} (baselines) or our \method (highlighted in \textcolor{darkgreen}{green}); all conditions share the same real fine-tuning data per task. In \textbf{\method-S}, no real demonstrations are used for training, so each column is a different source of task-specific training data: simulated point clouds versus our \method synthetic stereo. Across both, \method\ (\textbf{Ours}) gives the best performance.}
\vspace{-5pt}
\label{tab:real-world-results}
\end{table*}

\subsection{\method-S: Stereo Generation w/out a G.T. Left View}
\label{ssec:method2}

\method-L requires a ground-truth left video to warp. When no such video exists, we instead generate the left and right views in simulation and synthesize both directly, conditioning on a simulated trajectory, a language prompt, and two matching reference images of the real workspace, one for the left view and one for the right, that supply the target appearance. These reference images are easy to obtain as they only require capturing a single stereo photo of the robot scene.

\subsubsection{Generating Demonstrations from Simulation}
\label{ssec:trajectory}
We leverage RoboTwin~\cite{chen2026robotwin} to generate the demonstrations. Since the cameras can be placed anywhere in the scene, we position a second virtual camera at a chosen baseline from the first and render a true stereo pair for every frame. The disparity between the two views can be directly retrieved from the simulator's camera parameters and scene geometry so no depth estimation is needed and no disocclusions have to be filled. 
However, the simulation cannot provide realistic rendering since its textures, materials, and lighting differ substantially from the real world. We therefore convert each frame into a Canny edge map~\cite{canny1986edge}, as also done in CRAFT~\cite{chen2026craft}, which preserves the outlines of the arms and objects while removing the simulator's visual appearance. This leaves the video model free to synthesize a realistic scene while still following the simulated motion. We then tile the left and right frames into a single video frame so that one generation produces both views simultaneously. Because both views are denoised together, they remain consistent with one another.

\subsubsection{Generating the Synthetic Left and Right Videos}
\label{ssec:video-generation}
Let $\mathbf{V}^{c}$ denote the tiled Canny control video from Section~\ref{ssec:trajectory}, $I^{\text{ref}}$ the reference input, and $\ell$ the language prompt. Following CRAFT~\cite{chen2026craft}, we condition on real reference images of the workspace to guide appearance. Because we generate both views, $I^{\text{ref}}$ consists of two matching reference images, $I^{\text{ref}}_{L}$ and $I^{\text{ref}}_{R}$, captured as a single stereo photo of the workspace and tiled in the same layout as the generated frame. The language prompt $\ell$ is a short description of the scene and task, for example \textit{``two robot arms stacking two bowls on a wooden table.''} We use Wan2.2-Fun-A14B-Control~\cite{wan2025} as our
backbone and generate the tiled stereo video by sampling from
\begin{equation}
p_\phi(\hat{\mathbf{V}} \mid \mathbf{V}^{c}, I^{\text{ref}}, \ell),
\end{equation}
where $\hat{\mathbf{V}}$ is the generated video and $\phi$ denotes the model parameters. Unlike \method-L, which fine-tunes a Wan2.2 variant to inpaint the disoccluded regions of a warped view, here the pretrained model performs full video generation zero-shot, synthesizing the entire tiled stereo video from the conditioning inputs without any fine-tuning. The control video fixes
the structure and motion, the reference image provides the real-world appearance, and the language prompt
specifies the task. Splitting each generated frame gives the synthetic left and right videos $\hat{\mathbf{V}}^{L}$
and $\hat{\mathbf{V}}^{R}$ which form the stereo pair.

\begin{figure}[t]
    \centering
    \includegraphics[width=0.48\textwidth]{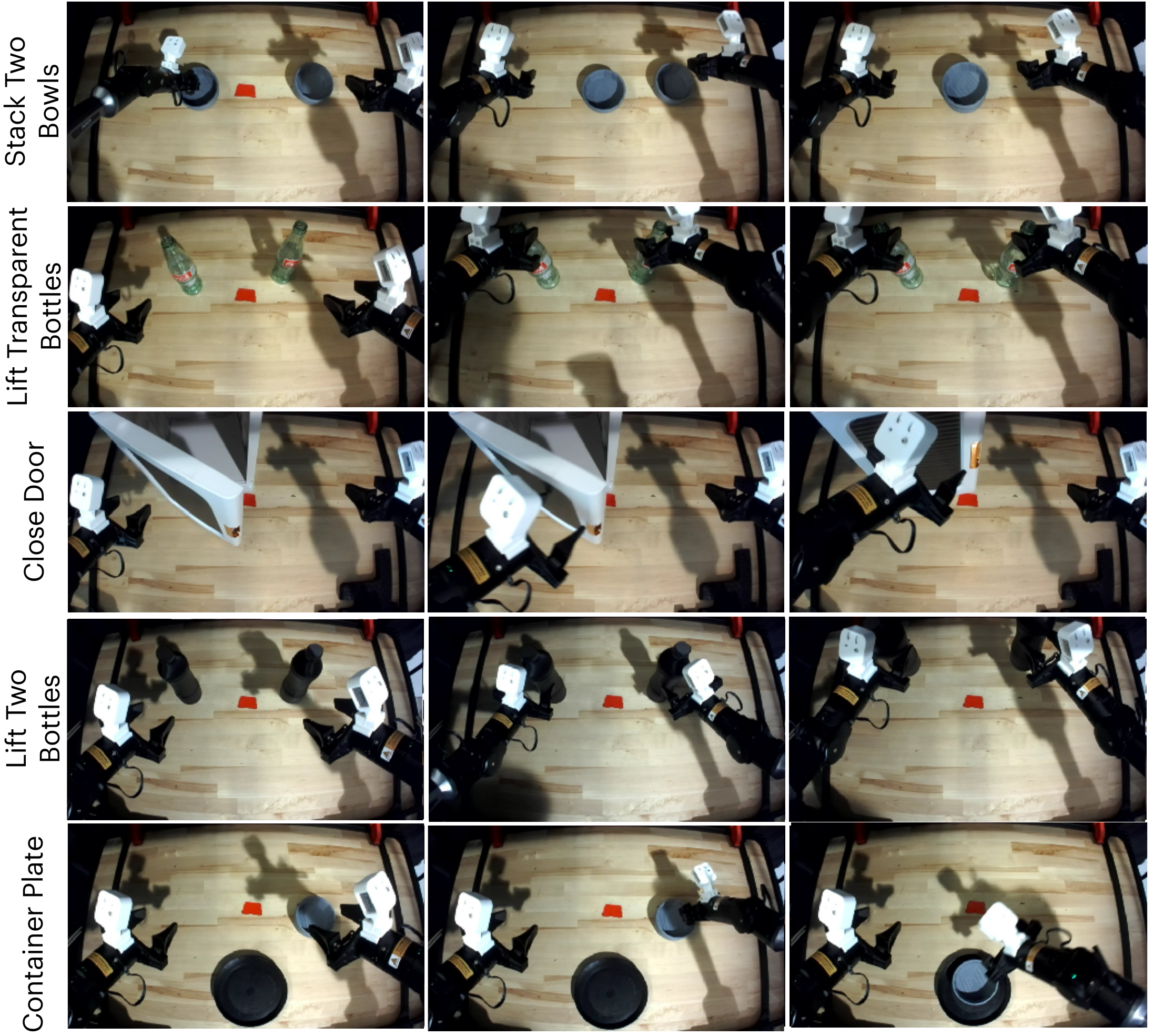}
    \caption{
        \textbf{Real-world rollouts.}
        Each row shows a successful policy rollout by ExStereo~\cite{liu2027exstereo} trained on \method data, from top to bottom: \texttt{\textbf{Stack Two Bowls}}, \texttt{\textbf{Lift Transparent Bottles}}, \texttt{\textbf{Close Door}}, \texttt{\textbf{Lift Two Bottles}}, and \texttt{\textbf{Container Plate}}. Both \method methods test \texttt{\textbf{Stack Two Bowls}}, while the other four tasks are tested on one of our methods. Columns show the progression of time (left to right).
    }
    \label{fig:rollouts}
    \vspace{-10pt}
\end{figure}

\section{Real-World Experiments}

\subsection{Real-World Experiment Setup}

We use two AgileX PiPER robot arms and a ZED Mini 2 camera. An experienced
roboticist used a Meta Quest 2 to collect teleoperation data. Because task
difficulty varies, we collect or generate a different number of demonstrations
for each task. For \method-L, the mid-training dataset consists of \num{3469} public monocular episodes lifted into stereo by \method-L and each baseline mid-trains on the same episodes converted by its own method. We then fine-tune on real teleoperated demonstrations captured with the ZED Mini 2: \num{200} for \texttt{\textbf{Stack Two Bowls}}, \num{50} for \texttt{\textbf{Lift Transparent Bottles}}, and \num{20} for \texttt{\textbf{Close Door}}. For \method-S, no real demonstrations are used and instead we generate synthetic stereo episodes per task from simulation: \num{200} for \texttt{\textbf{Stack Two Bowls}}, \num{50} for \texttt{\textbf{Lift Two Bottles}}, and \num{50} for \texttt{\textbf{Container Plate}}. We do not report simulation experiments, as our method generates stereo training data for real-world policy learning and is therefore evaluated directly on physical hardware.

We evaluate on three tasks for \method-L:
\begin{itemize}
\item \texttt{\textbf{Stack Two Bowls}}: two bowls must be stacked on top of each other in order, testing sequential manipulation.
\item \texttt{\textbf{Lift Transparent Bottles}}: two transparent glass bottles must be picked up from the table, testing perception of transparent objects.
\item \texttt{\textbf{Close Door}}: the left arm must close the door of a fridge, testing manipulation of an articulated object.
\end{itemize}

For \method-S, we reuse the \textbf{\texttt{Stack Two Bowls}} task from \method-L and evaluate on two extra tasks:
\begin{itemize}
\item \texttt{\textbf{Lift Two Bottles}}: two non-transparent bottles must be picked up from the table.
\item \texttt{\textbf{Container Plate}}: one bowl must be stacked on top of a plate.
\end{itemize}
We do not evaluate the \textbf{\texttt{Lift Transparent Bottles}} or \textbf{\texttt{Close Door}} tasks under \method-S. Articulated objects such as a fridge door require substantial per-object tuning to model in RoboTwin, and the difficulty of accurately simulating real-world contact dynamics makes their interactions hard to reproduce faithfully. In addition, current state-of-the-art video diffusion models still struggle to accurately reproduce transparent and highly specular surfaces. Here, we focus on evaluating our method under the current limitations of video generation models.
We compare \method-L against two strong monocular-to-stereo baselines, mid-training ExStereo, a stereo-conditioned policy built on $\pi_{0.5}$~\cite{intelligence2025pi05}, on each one's stereo data:

\begin{itemize}
    \item \textbf{No Mid.}: Skips mid-training and fine-tunes on the real demonstrations directly.
    \item \textbf{SVG~\cite{dai2025svg}}: a training-free method that uses Zeroscope to synthesize the right view.
    \item \textbf{StereoCrafter~\cite{zhao2024stereocrafter}}: a training-based method that fine-tunes Stable Video Diffusion on stereo pairs to synthesize the right view.
    \item \textbf{\method (Ours)}: we generate the synthetic stereo dataset with \method-L.
\end{itemize}
We do not benchmark against Geometric Reciprocity~\cite{lu2026grt}, as it is concurrent work with no publicly available code or weights at the time of publication. We also do not compare against other stereo-conditioned policies since ExStereo already outperforms existing methods at the time of writing. Although ExStereo~\cite{liu2027exstereo} is not publicly released at the time of publication, we obtained its implementation from its authors, who plan to release their polished version. We will also release our code to support reproducibility.

Since no existing method generates \textit{both} synthetic left and right views, we compare \method-S against a single baseline. In each case we train both ExStereo~\cite{liu2027exstereo} and StereoVLA~\cite{deng2026stereovlaenhancingvisionlanguageactionmodels}:
\begin{itemize}
    \item \textbf{Point Clouds}: we render ground-truth point clouds directly from
    the simulator and perturb every point with zero-mean isotropic Gaussian noise
    ($\sigma\!=\!\SI{4}{\milli\meter}$ per axis) to approximate real-sensor
    imperfections, then train the policies on this simulated 3D geometry using the same number of episodes as \method-S. 
    \item \textbf{\method (Ours)}: we generate the synthetic stereo dataset with \method-S, conditioning on the Canny edges and two reference images.
\end{itemize}
We do not evaluate StereoPolicy~\cite{han2026stereopolicyimprovingroboticmanipulation} as no code or public weights are available at the time of publication.

\begin{table}[t]
\centering
\small
\setlength{\tabcolsep}{5pt}
\renewcommand{\arraystretch}{1.15}
\begin{tabular}{lccc @{\hspace{12pt}} c}
\toprule
\textbf{Method} & PSNR$\uparrow$ & SSIM$\uparrow$ & LPIPS$\downarrow$ & CTC$\uparrow$ \\
\midrule
StereoCrafter~\cite{zhao2024stereocrafter} & 10.50 & 0.4574 & 0.6211 & 0.9807 \\
SVG~\cite{dai2025svg}                       & 12.83 & 0.5137 & 0.5128 & 0.9752 \\
\midrule
\method (Ours) & \textbf{15.31} & \textbf{0.5304} & \textbf{0.4696} & \textbf{0.9857} \\
\bottomrule
\end{tabular}
\caption{\textbf{Stereo generation quality for \method-L}, on 50 held-out real stereo clips. These clips are different from the demonstrations used for training.
Each method synthesizes a right view from the same ground-truth left video.
CTC measures temporal consistency across generated frames.
\method achieves the best score on every metric.
Here, $\uparrow$ means higher is better, and $\downarrow$ means lower is better.}
\label{tab:method1-quality}
\vspace{-8pt}
\end{table}

\subsection{Real-World Results}
Real-world results are presented in Table~\ref{tab:real-world-results} and qualitative rollouts are shown in Figure~\ref{fig:rollouts}. \method demonstrates substantially improved robustness across all tasks compared to baselines.
Aggregated over the three tasks, mid-training ExStereo on \method\ data succeeds on 50/60 trials (83.3\%) versus 36/60 (60.0\%) with no mid-training, with StereoCrafter in between at 40/60 (66.7\%). SVG performs worst at 31/60 (51.7\%), falling below no mid-training on every task. The largest gain is on \textbf{\texttt{Lift Transparent Bottles}}, where \method reaches 18/20 while no mid-training reaches only 11/20 and SVG only 10/20.

\subsubsection{\method-L Real-World Results}
\texttt{\textbf{Stack Two Bowls:}} The baselines frequently failed to grasp the bowls and closed the gripper beside or above them. With SVG mid-training the arms sometimes froze or cycled through the same motion for certain bowl placements. The baselines also often grasped the left bowl but missed the right one. Even after grasping both, they struggled to place one bowl cleanly on top of the other.

\texttt{\textbf{Lift Transparent Bottles:}} The baselines struggled to grasp the bottles because the most common failure was knocking a bottle over, since the opening to grasp is narrow and small depth errors drive the gripper into it.

\texttt{\textbf{Close Door:}} The baselines often misjudged the distance to the door and missed it, driving the arms into the fridge. The no mid-training baseline in particular reached too far to the right or too high and missed the door entirely. In a few cases the arm pushed with excessive force and tipped the fridge over, which we counted as a failure.

\subsubsection{\method-S Real-World Results}
Training the policy on ground-truth point clouds from the simulator does not transfer to the real world, as every task results in a 0\% success rate. We qualitatively observe that there is a substantial sim-to-real gap, because real stereo depth contains substantial structured sensor noise that clean simulated point clouds do not exhibit. Our point cloud baseline already perturbs each point with Gaussian noise ($\sigma\!=\!4\,\text{mm}$) and varying this noise did not yield any meaningful improvement because simple noise does not reproduce the material-dependent dropouts, missing regions, and edge artifacts of real depth sensors. This is consistent with prior work showing that naive noise augmentation is insufficient to close the sim-to-real gap for depth and point cloud policies~\cite{zhou2025diffudepgraspdiffusionbaseddepthnoise}.

\texttt{\textbf{Container Plate:}} Our method occasionally fails to grasp the right bowl, and even when it does, the bowl is sometimes not placed cleanly onto the plate.

\texttt{\textbf{Lift Two Bottles:}} The most common failure was knocking a bottle over, since the opening to grasp is narrow and small depth errors drive the gripper into it.

\begin{figure}[t]
    \centering
    \includegraphics[width=0.48\textwidth]{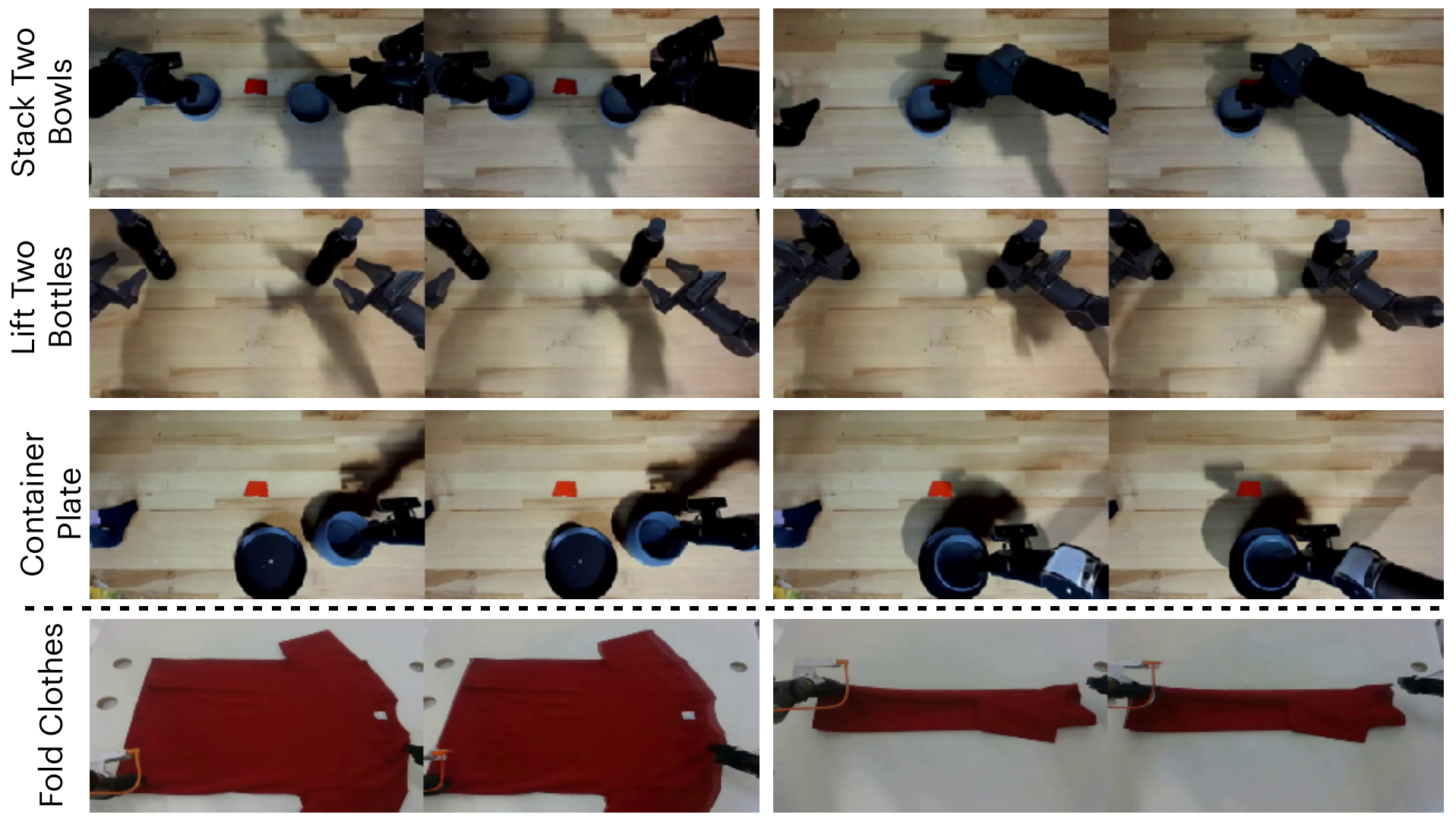}
    \caption{
        \textbf{Synthetic episode generation.}
        The top three rows show synthetic episodes generated by \method-S, one per task, from top to bottom: \texttt{\textbf{Stack Two Bowls}}, \texttt{\textbf{Lift Two Bottles}}, and \texttt{\textbf{Container Plate}}. Below the dashed line, the last row (\texttt{\textbf{Fold Clothes}}) shows a large-scale data generation example from \method-L, where a real monocular demonstration is lifted into stereo. Within each row, adjacent columns form a left--right stereo pair, and successive pairs show the progression of time from left to right.
    }
    \label{fig:synthetic_rollouts}
    \vspace{-10pt}
\end{figure}

\subsection{Generation Quality Analysis}
\textbf{Generation Quality.}
Since no stereo benchmark exists for robotic domains, we build our own evaluation set by collecting play data on our real stereo rig, giving paired real left and right videos to compare against. For \method-L, we run \method on the real left video and compare the generated right view to the true captured right view. As Table~\ref{tab:method1-quality} shows, \method beats both StereoCrafter and SVG on every metric---PSNR, SSIM~\cite{wang2004image}, LPIPS~\cite{zhang2018perceptual}, and CLIP Temporal Consistency (CTC)~\cite{radford2021clip}, which measures frame-to-frame CLIP feature similarity for temporal coherence, including a \num{2.48}\,dB PSNR gain over the strongest baseline. Absolute scores are lower than on standard stereo benchmarks because our ground truth is a real second camera with large disparities and imperfect rectification, unlike the perfectly rectified content those baselines are typically evaluated on.

\begin{figure}[t]
    \centering
    \includegraphics[width=0.48\textwidth]{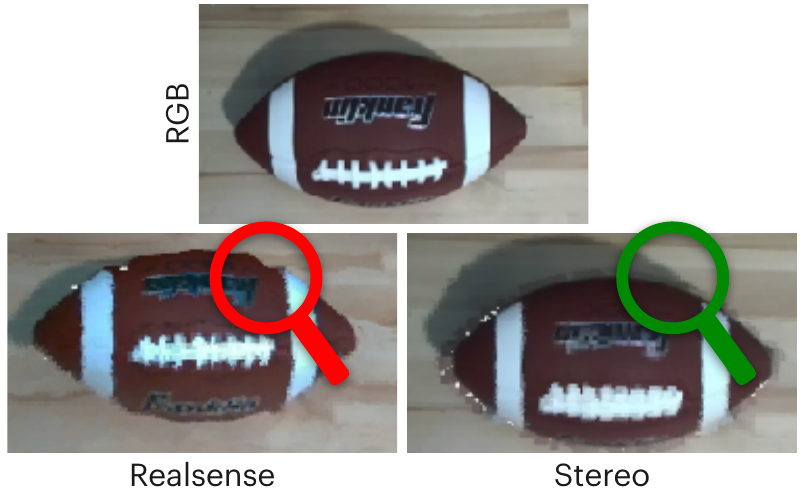}
    \caption{\textbf{Comparison of RealSense and stereo.} \textbf{Top:} an American football captured with a standard RGB camera. \textbf{Bottom left:} the 3D point cloud from an Intel RealSense D405 depth sensor where the \textcolor{red}{red} region shows geometric deformation of the ball's surface. \textbf{Bottom right:} the point cloud recovered by FoundationStereo~\cite{wen2025stereo} from a stereo pair captured with a ZED Mini 2 where the same \textcolor{darkgreen}{green} region shows minimal deformation of the object.}
    \label{fig:pointcloud_comparison}
    \vspace{-10pt}
\end{figure}

\section{Discussion}

As Figure~\ref{fig:pointcloud_comparison} shows, stereo recovers clean geometry on a specular object where an active depth sensor (Intel RealSense D405) deforms the surface, motivating the stereo data \method generates. \method provides two independent methods for producing stereo data, based on whether real monocular footage of the task exists. \textbf{\method-L} applies when a monocular dataset already exists, and we highlight two use cases: (1) mid-training a stereo policy on large-scale monocular data lifted into stereo, and (2) adapting an existing monocular dataset for a given task to a stereo policy. \textbf{\method-S} covers the analogous cases when no real monocular video is available: (1) building a large-scale mid-training dataset directly from a simulator, and (2) fine-tuning a stereo policy on a specific task by generating targeted stereo demonstrations.

\section{Conclusion}

We present \method, a video diffusion framework for synthesizing stereo data for bimanual manipulation without training on real stereo correspondences. Given an existing monocular demonstration, we warp the left view and inpaint the disoccluded regions to produce the corresponding right view. When no real video exists, we instead generate both views jointly from a simulated trajectory. Across real-world tasks on AgileX PiPER arms, \method data improve stereo-conditioned policies through both large-scale mid-training and task-specific fine-tuning, outperforming monocular-to-stereo baselines and simulated point clouds.
We hope this enables exciting research on stereo-conditioned foundation policies trained at scale.

\section{Acknowledgment}

Daniel Seita acknowledges generous support from Samsung Research America, NSF grant \#2447397, and the NVIDIA Academic Grant Program Award.

\bibliographystyle{IEEEtran}
\bibliography{main}

\clearpage
\appendix
\subsection{Additional Real-World Implementation Details}
\label{additionaldetails}
Our real-world setup uses a bimanual AgileX PiPER platform with a ZED Mini 2 stereo camera rigidly mounted to the frame. An expert operator collected the teleoperated demonstrations with a Meta Quest 2 for the tasks that require real data. For \texttt{\textbf{Stack Two Bowls}}, we place a small piece of red tape at the center of the workspace to mark where the bowls should be stacked.

We 3D-printed the objects for \texttt{\textbf{Stack Two Bowls}}, \texttt{\textbf{Lift Two Bottles}}, and \texttt{\textbf{Container Plate}} on a Bambu printer. For \texttt{\textbf{Lift Transparent Bottles}}, we used commercially available glass beverage bottles, and for \texttt{\textbf{Close Door}}, we used a commercially available articulated fridge toy.

\subsection{StereoEngine-L Implementation Details}
For the mid-training stage of \method and the baselines, we draw on a collection of publicly available monocular, bimanual PiPER datasets from HuggingFace. \method-L is not restricted to AgileX PiPER arms as the only requirement is that monocular data exist for the target embodiment. To apply \method-L to a new embodiment, we recommend scraping HuggingFace for publicly available datasets of that embodiment, or collecting play data directly.

We list the original sources below, grouped by uploader.

\begin{itemize}
  \item \textbf{csjmacmi}:
    \href{https://huggingface.co/datasets/csjmacmi/piperx_pick_and_place_v1}{piperx\_pick\_and\_place\_v1},
    \href{https://huggingface.co/datasets/csjmacmi/piperx_pick_and_place_v1_2}{v1\_2},
    \href{https://huggingface.co/datasets/csjmacmi/piperx_pick_and_place_v1_3}{v1\_3},
    \href{https://huggingface.co/datasets/csjmacmi/piperx_pick_and_place_v1_4}{v1\_4},
    \href{https://huggingface.co/datasets/csjmacmi/eval_piperx_pick_and_place_v1_4}{eval\_v1\_4},
    \href{https://huggingface.co/datasets/csjmacmi/eval_piperx_pick_and_place_v1_5}{eval\_v1\_5}
  \item \textbf{mjung11}:
    \href{https://huggingface.co/datasets/mjung11/nc_bi_piper_pot_car_apache_merged}{nc\_bi\_piper\_pot\_car\_apache\_merged},
    \href{https://huggingface.co/datasets/mjung11/nc_bi_piper_pp_grape_pot_depth_mj_A}{nc\_bi\_piper\_pp\_grape\_pot\_depth\_mj\_A},
    \href{https://huggingface.co/datasets/mjung11/nc_bi_piper_folding_mj_A}{nc\_bi\_piper\_folding\_mj\_A},
    \href{https://huggingface.co/datasets/mjung11/bi_piper_open_tissue_test}{bi\_piper\_open\_tissue\_test},
    \href{https://huggingface.co/datasets/mjung11/eval_bi_piper_tissue_xvla}{eval\_bi\_piper\_tissue\_xvla}
  \item \textbf{Autobrik}:
    \href{https://huggingface.co/datasets/Autobrik/bi_piper_pick_and_place_fruits_mantra}{bi\_piper\_pick\_and\_place\_fruits\_mantra}
  \item \textbf{gauravpradeep}:
    \href{https://huggingface.co/datasets/gauravpradeep/t01_piper_pick_and_place_bimanual_lerobot}{t01\_piper\_pick\_and\_place\_bimanual\_lerobot}
  \item \textbf{JohnnyXing}:
    \href{https://huggingface.co/datasets/JohnnyXing/001_2026041415_bipiper}{001\_2026041415\_bipiper}
  \item \textbf{moonyh1230}:
    \href{https://huggingface.co/datasets/moonyh1230/piper_wooden_block_dual_realsense_0922}{piper\_wooden\_block\_dual\_realsense\_0922}
  \item \textbf{RoboSynChallenge}:
    \href{https://huggingface.co/datasets/RoboSynChallenge/cobotmagic_Real_water_pouring}{cobotmagic\_Real\_water\_pouring}
  \item \textbf{RDLwicked}:
    \href{https://huggingface.co/datasets/RDLwicked/pick_up_green_block_drawer_piperX}{pick\_up\_green\_block\_drawer\_piperX}
  \item \textbf{Sohaib03}:
    \href{https://huggingface.co/datasets/Sohaib03/piperx_pick_and_place_v1_2}{piperx\_pick\_and\_place\_v1\_2}
  \item \textbf{12e21}:
    \href{https://huggingface.co/datasets/12e21/bi_piper_subset}{bi\_piper\_subset}
  \item \textbf{richardshkim}:
    \href{https://huggingface.co/datasets/richardshkim/bi_piper_dataset}{bi\_piper\_dataset}
\end{itemize}

\subsection{Limitations}
\method has several limitations:
\begin{enumerate}
    \item Video diffusion models struggle to synthesize transparent and highly reflective objects, which can degrade the quality of the generated stereo pairs.
    \item \method-L inherits the quality of the monocular depth estimator, so inaccurate depth yields geometrically inconsistent right views.
    \item \method-L's inpainting quality depends on its fine-tuning data as too little data gives weaker generations.
    \item \method-S is performed zero-shot but still requires careful per-scene prompt engineering and informative reference images, a limitation shared with CRAFT~\cite{chen2026craft}.
    \item \method-S relies on Canny edges from simulation, so larger camera distances produce noisier edges that can degrade the generated views.
\end{enumerate}

\begin{figure}[t]
    \centering
    \includegraphics[width=0.5\textwidth]{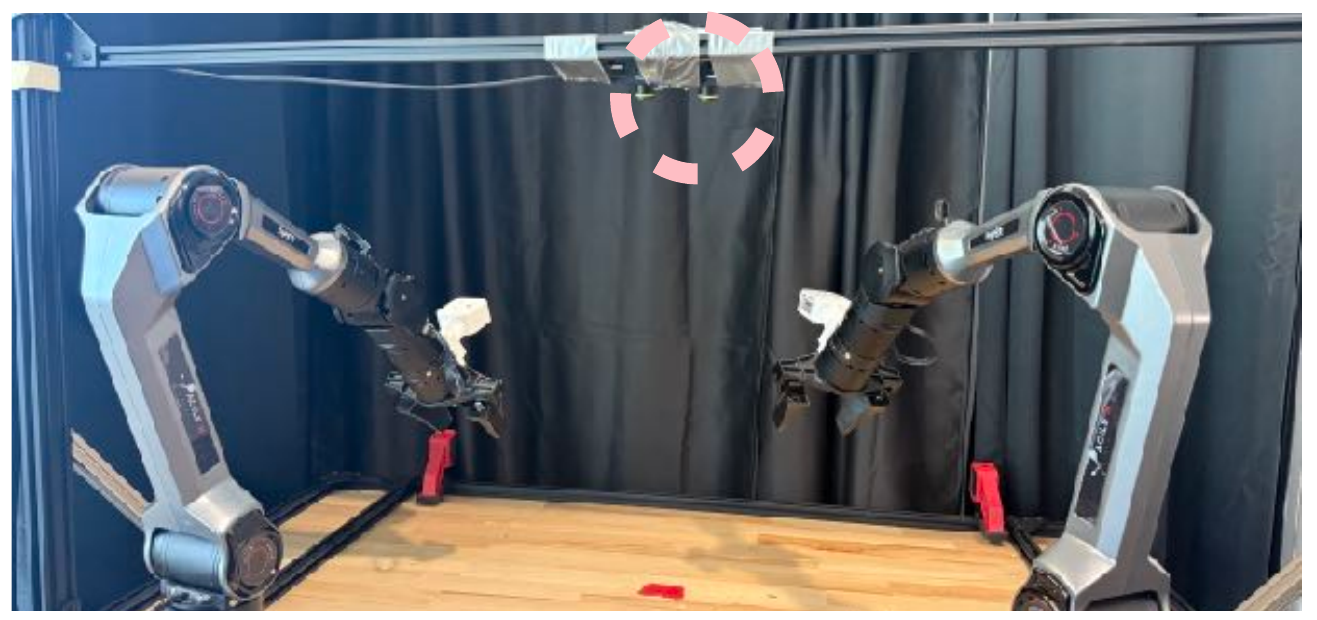}
    \caption{\textbf{Real-World Setup.} Our real-world platform uses two AgileX PiPER arms. The pink dashed circle marks the ZED Mini 2 stereo camera that captures the stereo pairs.}
    \label{fig:realworldsetup}
    \vspace*{-10pt}
\end{figure}

\end{document}